\documentclass[runningheads]{llncs}
\usepackage{graphicx}
\usepackage{amsmath}
\usepackage{booktabs} 
\usepackage{multirow}
\usepackage[margin=1.2in]{geometry}
\usepackage[round,sectionbib]{natbib}
\usepackage{xcolor} 
\usepackage{hyperref}
\hypersetup{
  colorlinks,
  citecolor=violet,
}
  
\begin{document}

\title{Zero-Shot Spam Email Classification Using \\Pre-trained Large Language Models}
\titlerunning{Zero-Shot LLM Spam Email Classification}

\author{Sergio Rojas-Galeano}
\authorrunning{S. Rojas-Galeano}

\institute{Universidad Distrital Francisco José de Caldas, Bogotá, Colombia\\
\email{srojas@udistrital.edu.co}}

\maketitle              

\begin{abstract}
This paper investigates the application of pre-trained large language models (LLMs) for spam email classification using zero-shot prompting. We evaluate the performance of both open-source (Flan-T5) and proprietary LLMs (ChatGPT, GPT-4) on the well-known \mbox{SpamAssassin} dataset. Two classification approaches are explored: (1) truncated raw content from email subject and body, and (2) classification based on summaries generated by ChatGPT. Our empirical analysis, leveraging the entire dataset for evaluation without further training, reveals promising results. Flan-T5 achieves a 90\% F1-score on the truncated content approach, while GPT-4 reaches a 95\% F1-score using summaries. While these initial findings on a single dataset suggest the potential for classification pipelines of LLM-based subtasks (e.g., summarisation and classification), further validation on diverse datasets is necessary. The high operational costs of proprietary models, coupled with the general inference costs of LLMs, could significantly hinder real-world deployment for spam filtering.

\keywords{Email spam classification, LLMs, Zero-shot prompting.}
\end{abstract}

\linespread{1.15}\selectfont
\setlength{\parskip}{3pt}

\section{Introduction}
The ever-evolving digital landscape continues to be plagued by unwanted emails, commonly known as spam. From the infamous Nigerian prince scams of the early internet era \citep{Okosun2023} to the increasingly sophisticated phishing attempts, spammers constantly adapt their tactics to circumvent traditional filtering methods \citep{Salman2024}. This ongoing arms race necessitates ever more intelligent and adaptable solutions for spam detection.

The challenge of filtering spam emails remains a subject of significant academic interest \citep{Dada2019}. This sustained focus reflects the substantial economic and social costs associated with spam. Studies estimate that spam not only disrupts productivity and incurs economic costs within organisations \citep{Rao2012} but also acts as a vector for malware distribution and financial fraud, ultimately disrupting communication networks \citep{Fonseca2016}.

Traditional spam filtering techniques have limitations. Rule-based filtering, while efficient, struggles to adapt to novel spamming techniques, requiring frequent updates. Statistical methods offer more flexibility but necessitate substantial training data and can be computationally expensive. Machine learning (ML) methods have achieved success in spam detection \citep{Deng2023}. However, their effectiveness is hampered by the evolving nature of spam. This drift in spam tactics (known as \textit{concept drift} \citep{Dada2019}) negatively impacts the ability of these techniques to generalise and perform well on unseen emails. 

Recent advancements in Large Language Models (LLMs) \citep{Hadi2023} offer a promising new approach to combat spam. These powerful AI models, trained on massive text and code datasets, have demonstrated remarkable capabilities in tasks ranging from natural language translation to text summarisation \citep{Kalyan2023}. Their ability to understand complex linguistic patterns and relationships opens up a compelling avenue for spam detection, where emails often rely on deceptive language and emotional triggers. In fact, LLMs have been shown to be well-suited for text classification tasks \citep{Fields2024}, where the goal is to assign a predefined label to a piece of text. The spam classification problem can be framed as a text classification task where the LLM receives an email as input and outputs a category label (e.g., spam or ham).

This paper contributes to this exciting field by exploring the potential of pre-trained LLMs for zero-shot classification of spam emails. We investigate the effectiveness of zero-shot prompting, where models are directly evaluated on the task without further fine-tuning. We compare the performance of both open-source and proprietary LLMs: ChatGPT \citep{Brown2020}, GPT-4 \citep{Openai2024}, and Flan-T5 \citep{Chung2022}. Below we outline our specific contributions:

\renewcommand{\labelitemi}{$\bullet$}
\begin{itemize}
    \item We study the capabilities of pre-trained LLMs for out-of-the-box spam email classification using zero-shot prompting.
    
    \item We propose two classification scenarios: (1) \textit{raw content}, where LLMs classify content truncated to a specific length from the email subject and body, and (2) \textit{summarized content}, where ChatGPT first summarises the email before LLMs classify the generated summary.
    
    \item Our empirical results demonstrate promising performance, achieving accuracy rates of 94\% (Flan-T5) and 97\% (GPT-4), respectively, for these two approaches.
\end{itemize}

\textbf{Related Work.} Machine learning has long been applied to email spam detection, with techniques such as Naive Bayes, logistic regression, decision trees, and Support Vector Machines achieving success \citep{Deng2023, Dada2019, Gangavarapu2020}. While these methods offer efficacy, they struggle to adapt to the ever-evolving tactics of spammers. As spam emails become more sophisticated and spam patterns shift over time due to evolving spammer tactics (known as concept drift), traditional approaches require constant updates or large, labeled datasets, which can be expensive and time-consuming to maintain. Furthermore, the dynamic nature of spam causes the data distribution to shift between training and testing phases, significantly impacting the generalisation performance of spam filters \citep{Janez2023}. Additionally, spam tactics often employ adversarial techniques, further complicating detection \citep{Rojas2017}.

Deep learning architectures like convolutional neural networks (CNNs) and long short-term memory (LSTM) networks have emerged as powerful tools for spam detection, offering superior performance in handling complex email content \citep{Roy2020, Malhotra2022}. However, these deep learning models require significantly more data and computational resources compared to conventional methods, which can be a limiting factor for practical deployment.

Large language models (LLMs) based on transformers offer a promising new direction for spam detection. These models, with their ability to learn complex relationships within text, have the potential to outperform traditional and deep learning methods. \cite{Yaseen2021} explored fine-tuning pre-trained transformer models like BERT \citep{Devlin2018} for spam filtering tasks. Other studies have shown success in applying BERT to counter adversarial tactics like substitution attacks in SMS spam detection compared to conventional n-gram based models \citep{Rojas2021}.

Furthermore, recent work explores fine-tuning LLMs like Flan-T5 and BERT for spam email classification \citep{Labonne2023}, demonstrating their effectiveness in the few-shot learning scenario even with limited training data. Interestingly enough, LLMs have been used for both sides of the phishing coin: detecting phishing emails, a special class of spam emails, but also creating them. They can not only identify these malicious messages, but also devise the content for a phishing email or a complete landing webpage according to \cite{Heiding2024} and \cite{Roy2023}.

\section{Methods}
The SpamAssassin public corpus\footnote{\url{https://spamassassin.apache.org/old/publiccorpus/}} is a widely used benchmark dataset for spam filtering tasks. Derived from various sources like public forums, direct correspondence, and news website newsletters, it comprises 6,047 messages with a roughly 31\% spam ratio. While the corpus categorizes messages into difficulty levels (e.g., spam, easy ham, hard ham), these categories are not relevant for our zero-shot approach. However, the SpamAssassin dataset's mixed sources and content diversity make it a valuable resource, providing a comprehensive and diverse range of messages for testing and evaluation of spam detection algorithms. For our research, the most recent iterations of these categories from 2003 and 2005 were utilised.

\subsection{Large Language Models}
As discussed earlier, LLMs are emerging as powerful tools for various text analysis tasks  \citep{Zhao2023, Hadi2023}. These models, built upon the Transformer architecture \citep{Vaswani2017}, are pre-trained on massive text datasets, enabling them to perform tasks like text classification without additional fine-tuning, a property referred to as few or zero-shot learning \citep{Brown2020}. This paper explores the effectiveness of three LLMs for zero-shot spam email classification:

\begin{itemize}
\setlength\itemsep{1em}
    \item \textbf{ChatGPT.} This LLM, developed by OpenAI, excels in tasks involving language comprehension and generation, particularly in conversational settings \cite{Kalyan2023}. Its focus on context and fluency, along with its training on diverse text corpora, makes it suitable for analyzing the often nuanced language used in spam emails. Its core model is an evolution of the GPT-3.5 architecture \citep{Brown2020}.

    \item \textbf{GPT-4.} Building on the success of ChatGPT, GPT-4 boasts deeper architectures and enhanced training strategies to improve contextual understanding and long-range dependency handling in text \citep{Openai2024}. This advancement can be beneficial for classifying spam emails that rely on complex phrasing or manipulate context. However, its increased complexity might introduce computational limitations for usability in resource-constrained environments \citep{Hadi2023}. Nonetheless, similar to other GPT models, GPT-4 necessitates careful evaluation of potential bias and privacy issues in training data for ethical considerations \cite{Hadi2023}.
    
    \item \textbf{FLAN-T5.} This LLM combines the strengths of Few-shot Language Adaptation Networks (FLAN) and the Text-to-Text Transfer Transformer (T5) \citep{Chung2022}, building upon instruction-tuned approaches \citep{Wei2021} and the robust text-to-text performance of T5 \citep{Lester2021}. Its ability to fine-tune on smaller datasets makes it a strong candidate for zero-shot classification tasks, potentially adapting well to the specific domain of spam emails even with limited labelled data.
\end{itemize}

In this work, we prioritise evaluating the effectiveness of these LLMs for zero-shot spam email classification using prompting, as explained in the next section. Their unique strengths are particularly relevant in this context.

\subsection{Zero-shot learning}
Zero-shot learning (ZSL) is a groundbreaking approach that enhances the capabilities of LLMs by enabling them to perform tasks on unseen categories \citep{Wei2021,Brown2020}. Unlike traditional supervised learning, which requires extensive labelled data for each task, ZSL allows LLMs to generalise their knowledge to entirely new domains with minimal or no additional training data. In the context of spam detection, this translates to classifying emails from unknown senders or containing novel tactics by leveraging semantic similarity to spam-triggering patterns encountered during pre-training on massive text corpora.

This approach is particularly attractive for spam detection as spam tactics constantly evolve. By identifying generalisable text-label relationships, ZSL models could potentially adapt to unseen spam formats even with limited training data. However, challenges remain, as previously mentioned. ZSL models might struggle with tasks requiring nuanced understanding or reasoning that cannot be captured solely through semantic relationships, and biases present in the pre-training data can be amplified in ZSL settings.

\subsection{Classification Scenarios}
\begin{figure}
    \centering
    \begin{tabular}{cc}
        \begin{minipage}{0.45\textwidth}
            \centering
            \includegraphics[width=\linewidth]{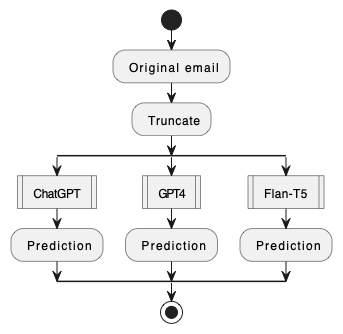} 
        \end{minipage}
        &
        \begin{minipage}{0.45\textwidth}
            \centering
            \includegraphics[width=\linewidth]{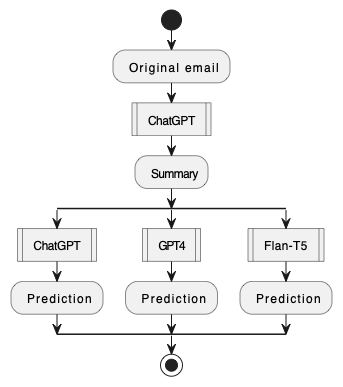} 
        \end{minipage} \\
        (a) & (b)
    \end{tabular}
    \caption{Schematic of two zero-shot LLM approaches for spam email classification: (a) Prediction from raw email content with truncation; (b) Prediction from a ChatGPT-generated email summary. Both approaches utilise three LLMs (ChatGPT, GPT-4, and FLAN-T5) for independent classification.}
    \label{fig:approaches}
\end{figure}

We devised two distinct classification scenarios to evaluate the performance of the zero-shot LLM spam email classification approach. Both scenarios leverage the inherent capabilities of LLMs for text processing and analysis. A schematic representation of these approaches is provided in Fig. \ref{fig:approaches} for easier visualisation.

\textbf{Prediction from Raw Content.} The first scenario directly utilises the truncated raw email content (both subject line and body text) for classification due to LLM input context limitations. This approach assesses the individual classification accuracy of each LLM in isolation, evaluating their ability to discriminate between spam and legitimate emails based solely on the original content. Formally, let $x_i = (s_i, b_i)$ represent the $i^{th}$ email instance, containing both subject line $s_i$ and body text $b_i$. Each LLM $M_j$ takes the truncated content $\tilde{x}_i = (s_i, \tilde{b}_i)$ as input and outputs a predicted class label $y_i^j \in \{\texttt{spam}, \texttt{ham}\}$. The performance of each $M_j$ is evaluated using standard classification metrics on the entire dataset.

\textbf{Prediction from Summary.} The second scenario employs a two-stage process. In the initial stage, ChatGPT acts as a summarising pre-processor, generating concise summaries $z_i$ of the email content, effectively condensing the information for each LLM. This summarised version captures the key points of the email and potentially eliminates irrelevant details. Subsequently, each LLM independently processes the summarised email content $z_i$ and provides a predicted class label $y_i^j$. Formally, the summarising pre-processor $S$ takes the raw email content $x_i$ as input and generates a summary $z_i = S(x_i)$. Each LLM $M_j$ then independently processes the summary $z_i$ and outputs a predicted class label $y_i^j$. This approach aims to improve classification accuracy by leveraging diverse perspectives from the model-generated summaries.

\subsection{Prompt design}
Prompt design plays a critical role in optimising LLMs for tasks like spam classification.  Acting as instructions and context, prompts guide the LLM's responses, shaping its final output. They bridge the raw input data with task objectives, enabling the LLM to understand the nuances of the task and generate accurate predictions \citep{Reynolds2021}. Effective prompt design considers task specificity, input complexity, and desired output format, ultimately enhancing LLM performance and adaptability. In our study, prompt design is crucial, as it guides LLMs through the spam classification task. The specific prompts devised for the two classification scenarios are presented in Table \ref{tab:prompts}.

\begin{table}[h!]
\centering
\caption{Prompts for zero-shot LLM spam email classification}
\label{tab:prompts}
\begin{tabular}{|@{\hspace{2mm}}p{0.12\linewidth}@{}|p{0.1\linewidth}|p{0.75\linewidth}|}
\hline
\multicolumn{1}{|c|}{\textbf{Approach}} & \multicolumn{1}{c|}{\textbf{Type}} & \multicolumn{1}{c|}{\textbf{Prompts}} \\ \hline\hline
\multirow{10}{2cm}{Prediction from Raw Content} 
& 
\multirow{2}{1.5cm}{ System} & \textit{You are an expert email spam classifier, capable of accurately identifying emails as either spam or ham based on their [SUBJECT] and [BODY].} \\
\cline{2-3}
& 
\multirow{7}{1.5cm}{ Task} & \textit{The text of an email will be delimited within triple hash symbols ('\#\#\#') and will consist of two sections: the subject and the body, which will be tagged accordingly as [SUBJECT] and [BODY]. Please classify the following email with a one-word answer, ``spam'' or ``ham''.} 

\#\#\#

\{email\}

\#\#\#

\textit{Answer:}
\\ \hline\hline
 
\multirow{14}{2cm}{Prediction from Summary\\ (Step 1)} 
& 
\multirow{7}{1.5cm}{ System} & \textit{Assume the role of an expert email summariser with the ability to accurately identify and analyse various elements of emails, including sender, receiver, subject, date, content type, content, priority, email server, email protocol, and more. You should also be capable of recognising elements within the email's content, such as its purpose, intent, style, vocabulary, languages (including natural and computer languages), and be able to provide a comprehensive description and an accurate summary, encompassing the aforementioned elements and aspects.} \\
\cline{2-3}
& 
\multirow{7}{1.5cm}{ Task} & \textit{The content of an email will be provided below, enclosed within triple hash symbols ('\#\#\#'). Please summarise it ensuring that your entire answer stays within 500 tokens.} 

\#\#\#

\{email\}

\#\#\#

\textit{Answer:}
\\ \hline

\multirow{8}{2cm}{Prediction from \\  Summary\\  (Step 2)} 
& 
\multirow{3}{1.5cm}{ System} & \textit{Assume the role of an expert email spam classifier, capable of accurately identifying emails as either spam or ham based on a summary of their most relevant elements.} \\ \cline{2-3}
& 
\multirow{7}{1.5cm}{ Task} & \textit{The summary of an email will be provided below, enclosed within triple hash symbols ('\#\#\#'). Please classify it with a one-word answer, ``spam'' or ``ham''.} 

\#\#\#

\{summary\}

\#\#\#

\textit{Answer:} 
\\ \hline

\end{tabular}
\end{table}

\subsection{Performance Metrics}
To comprehensively evaluate the performance of the LLMs in both classification scenarios, we employed a variety of metrics that assess their classification accuracy and effectiveness \citep{Tharwat2020}. These metrics provide insights into the models' ability to distinguish between spam and legitimate emails. Table \ref{tab:metrics} provides concise definitions of the utilised metrics, facilitating a clear understanding of their application in this context.

\begin{table}
\caption{Classification performance metrics}
\label{tab:metrics}
\centering
\begin{tabular}{ccp{8cm}}
\hline
\textbf{Metric} & \textbf{Definition} & \multicolumn{1}{c}{\textbf{Description}} \\ \hline
\multirow{2}{2cm}{\centering Accuracy ($AC$)} & \multirow{2}{3cm}{\centering $AC = \frac{TP+TN}{P+N}$} & A measure of the overall correctness of the model's predictions \\ \hline
\multirow{3}{2cm}{\centering Balanced Accuracy ($BA$)}  & \multirow{3}{3cm}{$BA = \frac{1}{2}\left({\frac{TP}{P} + \frac{TN}{N}}\right)$} & A measure of the model's performance on both positive and negative instances, taking into account the imbalance in the dataset \\ \hline
\multirow{2}{2cm}{\centering Precision ($PR$)} & \multirow{2}{3cm}{\centering $PR = \frac{TP}{TP+FP}$} & A measure of how often the model correctly classifies an instance as positive \\ \hline
\multirow{2}{1.5cm}{\centering Recall ($RE$)} & \multirow{2}{3cm}{\centering $RE = \frac{TP}{TP+FN}$} & A measure of how often the model correctly identifies all positive instances \\ \hline
\multirow{4}{1.5cm}{\centering F1 score ($F1$)} & \multirow{4}{3cm}{\centering $F1 = \frac{2 PR \times RE}{PR + RE}$} & A balanced measure of precision and recall that considers both the model's ability to correctly identify positive instances and its ability to avoid false positives \\ 
\hline
\multicolumn{3}{c}{\scriptsize $P, N$: Total positives, negatives; $TP, TN$: True positives, negatives; $FP, FN$: False positives, negatives} \\
\end{tabular}
\end{table}

\section{Results}

\subsection{Prediction from Raw Content} 
Table \ref{tab:truncated} displays the performance metrics of the three LLMs (ChatGPT, GPT-4, and FLAN-T5) in the scenario of prediction from raw, truncated content. Analysing accuracy (AC), Flan-T5 achieves the highest score (94\%), outperforming GPT-4 (91\%) and ChatGPT (82\%). However, considering balanced accuracy (BA), GPT-4 and FLAN-T5 perform almost equally well (93\%), indicating similar capabilities in identifying both spam and legitimate emails despite potential class imbalances.

Further examination of precision (PR) and recall (RE) provides insights into the trade-offs for each LLM. GPT-4 exhibits a bias towards high recall (98\%), effectively capturing most spam emails. However, this performance is diminished by misclassifications of legitimate emails, resulting in a much lower precision (79\%). In contrast, FLAN-T5 achieves a better balance between precision (90\%) and recall (89\%), indicating its suitability for accurately identifying both spam and legitimate emails while minimising misclassifications. Meanwhile, although ChatGPT excels at spam detection (97\% recall), its overall performance is impacted by its lowest precision (65\%).

Additionally, the F1-score (F1) corroborates these findings, with Flan-T5 emerging as the leading LLM with a well-rounded performance (90\%), followed by GPT-4 (88\%) and ChatGPT (78\%). These results suggest Flan-T5's effectiveness in zero-shot classification of spam emails based on raw, truncated content, without requiring further pre-processing or training. We note that these results are contingent on the truncation strategy employed and the characteristics of the spam email dataset.

\begin{table}[t]
    \centering
    
    \caption{Results on prediction from truncated content} \label{tab:truncated} 
    \begin{tabular}{llccccc}
    \toprule
    \textbf{Model} & &  \textbf{$AC$} &  \textbf{$BA$} &  \textbf{$PR$} &    \textbf{$RE$} &  \textbf{$F_1$} \\
    \midrule
    ChatGPT & &  0.823015 &       0.862831 &   0.645353 &  0.970063 &  0.775073 \\
    GPT-4 & &  0.914314 &       0.932680 &   0.794055 &  0.982143 &  0.878140 \\
    Flan-T5 & &  0.936932 &       0.925283 &   0.904357 &  0.893908 &  0.899102 \\
    \bottomrule
    \end{tabular} 

\bigskip\bigskip

    \caption{Results on prediction from summary} \label{tab:summary}
    \begin{tabular}{llccccc}
    \toprule
    \textbf{Model} & &  \textbf{$AC$} &  \textbf{$BA$} &  \textbf{$PR$} &    \textbf{$RE$} &  \textbf{$F_1$} \\\midrule
    ChatGPT & &  0.933488 &       0.937236 &   0.856190 &  0.947313 &  0.899450 \\
    GPT-4 & &  0.968729 &       0.955496 &   0.979248 &  0.919916 &  0.948655 \\
    Flan-T5 & &  0.928690 &       0.889745 &   0.984798 &  0.785037 &  0.873644 \\
    \bottomrule
    \end{tabular}
\end{table}

\subsection{Prediction from Summary}

Analysing Table~\ref{tab:summary}, a general trend towards improved performance is observed for most LLMs when using email summaries for spam classification, compared to raw content. GPT-4 emerges as the front-runner with the highest overall accuracy (97\%), followed by ChatGPT (93\%). Flan-T5 maintains a similar accuracy score to the previous scenario (93\%).

The positive effect of summarisation pre-processing is more evident in the precision and recall results. GPT-4 shows a significant improvement (PR: 98\%, RE: 92\%), yielding a high F1-score of 95\%. This improvement can potentially be attributed to the summaries effectively condensing the email content and highlighting key spam-indicative phrases, which GPT-4 excels at capturing. ChatGPT also benefits from summaries (PR: 86\%, RE: 95\%), achieving a high recall but falling short of GPT-4's ability to capture all spam emails. However, the increase in precision indicates fewer misclassifications compared to raw content, reflected in its F1-score (90\%).

Flan-T5 exhibits exceptional precision (98\%), rarely misclassifying legitimate emails when using summaries. However, its struggle with recall (79\%) suggests difficulty in identifying some spam emails based solely on condensed information. This imbalance is reflected in its F1-score (87\%), lower than the other models despite its high precision. The balanced accuracy (BA) metric ranks GPT-4 first (96\%), followed by ChatGPT (94\%), and lastly, Flan-T5 (89\%). Overall, feeding email summaries rather than raw content seems to positively impact GPT models' performance, likely due to the focused content highlighting spam-related cues. While Flan-T5 increases its precision, further investigation is needed to address its recall limitation in this summarisation scenario.

\section{Discussion}
Our empirical results demonstrate the potential of pre-trained large language models (LLMs) like ChatGPT, GPT-4, and Flan-T5 for zero-shot spam email classification. These models achieved strong performance on both raw content and email summaries, showcasing their versatility and ability to generalise from existing knowledge without fine-tuning. This eliminates the need for task-specific training data (and labelling effort), making the approach efficient, scalable, and particularly suited to dynamic environments with frequent new spam tactics. For instance, Flan-T5 achieved a well-rounded performance (90\% F1-score) on raw content classification, demonstrating its capability in handling unseen spam patterns without additional training.

The recent surge in LLM research for spam detection, as evidenced by studies like \cite{Heiding2024, Koide2024, Wu2024}, and \cite{Labonne2023}, highlights the effectiveness of LLMs in tackling this persistent problem. While fine-tuned models like Spam-T5 in \cite{Labonne2023} can achieve higher accuracy (e.g., 97\% F1 score), they require extensive task-specific training data and incur significant training costs during fine-tuning. Our zero-shot approach offers a balance between performance and efficiency. While inference time can be high due to the need for GPU-based servers, this is outweighed by the benefits of avoiding continual fine-tuning with each new LLM release. This significantly reduces training and maintenance efforts, including financial costs.

\cite{Wu2024} evaluated the performance of in-context ChatGPT without fine-tuning on an English spam email dataset, achieving an F1 metric of 80\% with a 5-shot and enhanced prompt engineering strategy. Their study suggests that prompt optimisation can further improve zero-shot LLM performance for spam detection. Additionally, \cite{Koide2024} focused on phishing email classification, a specific type of spam, demonstrating that GPT-4 achieves a remarkable 99\% accuracy on a custom honeypot dataset using a sophisticated pipeline. While our approach utilises a simpler pre-processing step with summarisation, exploring those additional techniques like chain-of-thought inference and few-shot prompting could potentially improve detection accuracy, especially for complex spam tactics.

One limitation of our study is that we used a single summarisation approach (ChatGPT-generated summaries). Investigating the impact of different summarisation techniques on LLM performance in zero-shot spam classification is an interesting avenue for future work. Ultimately, this line of research holds significant promise for developing practical and adaptable spam filtering solutions.

\section{Conclusion}
This research contributes to the field of spam email classification by demonstrating the efficacy of pre-trained LLMs for zero-shot classification. The study's findings reveal that these LLMs, including open-source and proprietary models, can achieve high F1 scores without fine-tuning when classifying spam emails. The summarisation pre-processing approach proved particularly effective, achieving F1-scores comparable to state-of-the-art results obtained using fine-tuned LLMs. This underscores the potential of leveraging pre-trained LLMs for efficient and accurate spam email classification without extensive retraining efforts, offering a promising avenue for combating spam in digital communication channels. Its key strengths lie in eliminating the need for collecting and labelling task-specific training data and avoiding the costs of fine-tuning every time a new LLM version emerges. However, the main disadvantages are related to inference time and hardware costs due to the high computational demands of these models.

Future research can explore several interesting avenues. Firstly, hybrid approaches that combine zero-shot capability with selective fine-tuning could balance efficiency and performance. Secondly, investigating the impact of different truncation techniques and additional pre-processing steps on LLM classification accuracy is warranted. Thirdly, integrating domain-specific knowledge or pre-training LLMs on dedicated spam email datasets, especially in scenarios with limited information, has potential advantages. Finally, exploring the benefits of aggregating predictions from an ensemble of models holds promise.

An additional avenue of research could involve leveraging our summarisation-based zero-shot LLM approach for spam email classification to other spam-related scams, such as phishing website detection. This aligns with the results achieved by \cite{Koide2023} who applied ChatGPT for similar purposes. Their work, demonstrating high accuracy in identifying phishing websites without extensive training data, suggests a potential path forward for a unified LLM-based approach to combat both spam emails and malicious websites.


\renewcommand{\bibname}{References}
\bibliographystyle{apalike}
\footnotesize
\bibliography{biblio}

\end{document}